\documentclass{article}
\usepackage{graphicx} % Required for inserting images
\usepackage{booktabs} % For \toprule, \midrule, \bottomrule in tables
\usepackage{amsmath, amssymb, url, verbatim}
\usepackage{tabularx}
\usepackage{array} % Required for the C column type
\usepackage{rotating}
\usepackage[margin=1in]{geometry}
\usepackage{pifont} % For \xmark and \cmark
\usepackage{pifont} % For \xmark and \cmark
\newcommand{\xmark}{\ding{55}} % ✗

\title{\textbf{RLHF: A comprehensive Survey for Cultural, Multimodal and Low Latency Alignment Methods}}
\author{%
  Raghav~Sharma\textsuperscript{1*}\\
  \texttt{sharma.raghav103@gmail.com}
  \and
  Manan~Mehta\textsuperscript{2}\\
  \texttt{manan.mehta2@gmail.com}
  \and
  Sai~Tiger~Raina\textsuperscript{2}\\
  \texttt{sai.raina@gmail.com}
}
\date{}

\newcolumntype{C}{>{\centering\arraybackslash}X}

\begin{document}

\maketitle

\begin{center}
\textsuperscript{1}\,Northeastern University, Boston, MA 02115, USA\\
\textsuperscript{2}\,University of Southern California, Los Angeles, CA 90089, USA
\end{center}

\section{Abstract}
Reinforcement Learning from Human Feedback (RLHF) is the standard for aligning Large Language Models (LLMs), yet recent progress has moved beyond canonical text-based methods. This survey synthesizes the new frontier of alignment research by addressing critical gaps in multi-modal alignment, cultural fairness, and low-latency optimization. To systematically explore these domains, we first review foundational algorithms, including PPO, DPO, and GRPO, before presenting a detailed analysis of the latest innovations. By providing a comparative synthesis of these techniques and outlining open challenges, this work serves as an essential roadmap for researchers building more robust, efficient, and equitable AI systems.

\section{Introduction}

The advent of models powered by Reinforcement Learning from Human Feedback (RLHF) marked a pivotal moment in artificial intelligence. This paradigm shift transformed Large Language Models from mere text generators into interactive, seemingly helpful assistants, making advanced AI accessible on a global scale. However, this initial success has also illuminated the limitations of a one-size-fits-all alignment strategy, revealing critical gaps in areas that previous surveys have not adequately addressed: \textbf{multi-modal alignment, cultural and demographic fairness, and methods for optimizing latency and cost.}
\\\\
To provide a clear guide to this new frontier, this survey offers a comprehensive and structured overview of these recent advances. We first establish the foundational alignment toolkit, covering the principles of RLHF and the three primary policy optimization techniques that form its backbone: Proximal Policy Optimization (PPO), Direct Preference Optimization (DPO), and Group Relative Policy Optimization (GRPO). We then present a systematic review of the latest methods designed to tackle the identified gaps, culminating in a comparative synthesis and an examination of open challenges. This work provides researchers with a complete roadmap to both the current state of the field and its most promising future directions.
\subsection{Reinforcement Learning in a Nutshell}
\label{sec:rl_basics}

The goal of Reinforcement Learning (RL) is to teach an agent to make good decisions by interacting with an environment and receiving feedback in the form of rewards. Common definitions used in RL:

\begin{itemize}
  \item \textbf{Agent}: the decision-making entity.
  \item \textbf{Environment}: everything the agent interacts with.
  \item \textbf{State} $s_t$: what the agent observes at time~$t$.
  \item \textbf{Action} $a_t$: the choice the agent makes at~$t$.
  \item \textbf{Reward} $r_t$: a scalar signal indicating the quality of $a_t$.
  \item \textbf{Policy} $\pi_\theta$: a parameterised distribution $\pi_\theta(a\mid s)$ that indicates agent’s strategy for picking actions.
\end{itemize}

The goal of the agent is to maximize its total rewards over time, which is often written as:
\begin{equation}
J(\pi_\theta)=\mathbb{E}_{\tau\sim\pi_\theta}\left[\sum_{t=0}^{\infty}\gamma^{t}\,r(s_t,a_t)\right]
  \label{eq:expected_return}
\end{equation}
where $\gamma\in[0,1)$ trades off immediate and future rewards and $\tau=(s_0,a_0,s_1,\dots)$ denotes a trajectory generated by $\pi_\theta$.

\subsection{Reinforcement Learning for Language Models}
\label{sec:rl_lm}

Large-language models (LLMs) trained purely by next-token prediction can write fluent text, but they do not automatically act in line with human preferences. Reinforcement Learning can help align LLMs to produce text that follows human stylistic and behavioral preferences. RL definitions can be adapted for LLMs as follows:

\begin{itemize}
  \item \textbf{Agent}: The language model.
  \item \textbf{Environment}: The user or a simulated evaluator.
  \item \textbf{State} $s_t$: The user prompt (and dialogue history).
  \item \textbf{Action} $a_t$: The entire completion (sequence of tokens).
  \item \textbf{Reward} $r_t$: Scalar score derived from human preference.
  \item \textbf{Policy} $\pi_\theta$: Next-token distribution of the LLM.
\end{itemize}

\subsection{Reinforcement Learning from Human Feedback (RLHF)}
\label{sec:rlhf}

Reinforcement Learning from Human Feedback (RLHF) is a technique used to integrate human preferences into AI systems. RLHF involves learning a reward model based on human judgments and optimizing the language model accordingly. This approach became popular after the release of ChatGPT, showcasing RLHF’s ability to produce safer, helpful and contextually appropriate responses. The basic pipeline of RLHF has 3 main steps:
\begin{enumerate}
  \item \textbf{Supervised Fine-Tuning (SFT)}: the base model is trained on a high-quality prompt-completion pairs dataset.
  \item \textbf{Reward-Model Training}: annotators rank different model outputs; and a separate network learns to predict these preferences, providing a scalar reward $r(x,y)\in\mathbb{R}$ for any prompt--completion pair.
  \item \textbf{Policy Optimisation}: Core to RLHF — the stage that governs how LLM weights are optimized to maximize the reward (learned in the previous step) while staying near the SFT reference model.
\end{enumerate}

\subsection{Policy-Optimisation Algorithms}
\label{sec:policy_optim}

The core of RLHF is policy optimisation: it aims to find the
parameters \( \theta \) of a policy \( \pi_{\theta} \) that maximise the expected cumulative reward.

\subsubsection{The Optimization Objective}

\begin{equation}
  J(\theta) \;=\; \mathbb{E}_{\tau \sim \pi_\theta}\!\bigl[\,R(\tau)\bigr]
\end{equation}

\noindent where
\begin{itemize}
  \item $\pi_\theta$ is the policy being optimized;
  \item $\tau=(s_0,a_0,s_1,a_1,\dots)$ is a trajectory generated by~$\pi_\theta$;
  \item $R(\tau)=\sum_{t=0}^T \gamma^{t}\,r(s_t,a_t)$ is the (possibly discounted) return with $\gamma\in[0,1)$;
  \item $\mathbb{E}_{\tau\sim\pi_\theta}[\cdot]$ denotes the expectation over trajectories induced by~$\pi_\theta$.
\end{itemize}

\subsubsection{The Policy Gradient Estimator}
To optimize $J(\theta)$, via gradient ascent, we use Policy Gradient Theorem. The gradient is estimated as:

\begin{equation}
  \nabla_\theta J(\theta)
  \;=\;
  \widehat{\mathbb{E}}_t\!\Bigl[
     \underbrace{\nabla_\theta \log \pi_\theta(a_t \mid s_t)}_{\text{score function}}\;
     \underbrace{\widehat{A}^{\pi}(s_t,a_t)}_{\text{advantage estimate}}\Bigr]
\end{equation}
where
\begin{itemize}
  \item \textbf{Score function} $\bigl(\nabla_\theta\log\pi_\theta(a_t\mid s_t)\bigr)$
     — the gradient of the selected action’s log-likelihood; it shows how a small change in~$\theta$ would increase or decrease the probability of choosing that action.

  \item \textbf{Advantage estimate} $\bigl(\widehat{A}^{\pi}(s_t,a_t)\bigr)$ — it measures how much better or worse the action performed relative to the policy’s baseline.
Usually
\[
       \widehat{A}^{\pi}(s_{t},a_{t})=
       Q^{\pi}(s_{t},a_{t})-V^{\pi}(s_{t}).
      \]
\end{itemize}
\begin{itemize}
  \item $V^{\pi}(s)$ --- \emph{state value}: expected return starting in $s$ and following $\pi$.
  \item $Q^{\pi}(s,a)$ --- \emph{action value}: expected return taking $a$ in $s$ then following $\pi$.
  \item $A^{\pi}(s,a)=Q^{\pi}(s,a)-V^{\pi}(s)$ --- \emph{advantage}: how much better or worse action $a$ is than the average action at $s$.
\end{itemize}

\bigskip
\noindent
This estimator, while foundational, is known to suffer from high variance, which can lead to unstable and inefficient training.

\subsubsection{Proximal Policy Optimisation (PPO)}
To mitigate the instability caused by large, unconstrained policy updates from the vanilla policy gradient estimator \cite{williams1992simple}, PPO was introduced \cite{schulman2017ppo}. PPO optimizes a surrogate objective function that restricts the extent of the policy update at each iteration.
\par\noindent
The PPO objective function is
\[
  L^{\text{CLIP}}(\theta)
  = \widehat{\mathbb{E}}_{t}\!\Bigl[
     \min\!\bigl(
       r_t(\theta)\,\widehat{A}_t,\;
       \operatorname{clip}\bigl(r_t(\theta),\,1-\epsilon,\,1+\epsilon\bigr)\,\widehat{A}_t
     \bigr)
    \Bigr]
\]
where,
\begin{itemize}
  \item \textbf{Probability Ratio}
       \[
         r_t(\theta)=
         \frac{\pi_\theta(a_t\mid s_t)}
            {\pi_{\theta_{\mathrm{old}}}(a_t\mid s_t)},
       \]
       comparing the likelihood of an action under the current and previous policies.
  \item \textbf{Clipping Mechanism:} the \texttt{clip} function constrains $r_t(\theta)$ to the trust region $[\,1-\epsilon,1+\epsilon\,]$.
  \item \textbf{Minimum Operator:} taking the minimum of the unclipped and clipped terms yields a conservative bound on the policy update.
\end{itemize}

\medskip
\noindent
\textbf{Result.} The objective penalizes large policy changes that push $r_t(\theta)$ outside the trust region, thereby enhancing stability.

\subsubsection{Group Relative Policy Optimisation (GRPO)}
While PPO enhances stability, it relies on a separately trained state--value function $V^{\pi}(s)$ for advantage estimation. This introduces significant computational overhead, as we are simultaneously training (policy, reward and value networks). GRPO dispenses the need for a separate value network.
For each prompt~$x$, the current policy produces a group of $G$ candidate
responses $\{y_i\}_{i=1}^{G}$, each scored by a reward signal
$r_i = R_{\phi}(y_i, x)$.
The advantage is defined as the normalised deviation of each reward from the group statistics.
\begin{equation}
\mu     = \frac{1}{G}\sum_{j=1}^{G} r_j,
\qquad
\sigma  = \sqrt{\frac{1}{G}\sum_{j=1}^{G}\bigl(r_j - \mu\bigr)^{2}},
\qquad
A_i     = \frac{r_i - \mu}{\sigma}.
\label{eq:grpo_advantage}
\end{equation}

This empirical baseline replaces $V^{\pi}(s)$ and is plugged directly into $L^{\text{CLIP}}$, eliminating the need for a separate value--function network.

\subsubsection{Direct Preference Optimisation (DPO)}
DPO reframes the RLHF alignment problem \cite{christiano2017deep} into a straightforward classification task \cite{rafailov2023direct}. It bypasses the traditional RL pipeline of explicit reward modeling and policy optimization \cite{ouyang2022training}. It directly fine-tunes the policy on a static set of human-ranked answers, turning the whole pipeline into one offline loss minimisation.
Given a static dataset of human preferences,
\[
  \mathcal{D} \;=\; \{(x,\,y_w,\,y_l)\},
\]
where \(y_w\) is the preferred and \(y_l\) the dispreferred response to
prompt \(x\).

\paragraph{DPO loss function}
\begin{equation}
  L_{\text{DPO}}\!\bigl(\pi_\theta;\,\pi_{\text{ref}}\bigr)
  \;=\;
  -\,\mathbb{E}_{(x,y_w,y_l)\sim\mathcal{D}}
  \!\left[
    \log \sigma\!\Bigl(
      \beta \log\!\frac{\pi_\theta(y_w\mid x)}{\pi_{\text{ref}}(y_w\mid x)}
      -
      \beta \log\!\frac{\pi_\theta(y_l\mid x)}{\pi_{\text{ref}}(y_l\mid x)}
    \Bigr)
  \right],
  \label{eq:dpo-loss}
\end{equation}
where
\begin{itemize}
  \item \(\pi_{\text{ref}}\) is a fixed \textbf{reference policy} (typically the initial,
       pre-trained model);
  \item \(\beta > 0\) is a \textbf{temperature} that controls deviation from the reference;
  \item \(\sigma(z) = \tfrac{1}{1 + e^{-z}}\) is the \textbf{sigmoid} function.
\end{itemize}

Equation~\eqref{eq:dpo-loss} directly increases the relative log-probability of the preferred response over the dispreferred one, weighted by their implicit reward difference.
Consequently, DPO optimises the policy directly from preference data, removing the need for a separate reward model and the instability of RL.

\section{Survey Methodology}

This survey synthesises recent advances in language model alignment based on a systematic review protocol. Candidate papers were selected based on the following criteria:

\begin{itemize}
    \item \textbf{Search Scope:} We reviewed papers from 2023--2025 across arXiv (cs.CL, cs.LG, cs.CV) and major AI/NLP conferences (e.g., NeurIPS, ICML, ICLR, ACL), focusing on text and multi-modal alignment.

    \item \textbf{Inclusion Criteria:} Included methods must perform reinforcement learning or explicit preference optimisation beyond supervised fine-tuning. We excluded works limited to prompt engineering or static filters unless integrated into an RL framework.

    \item \textbf{Analytical Framework:} Each included method was analysed along four axes: (1) \textbf{Reward Source} (human, synthetic, or self-improving); (2) \textbf{Optimisation Style} (policy gradient, preference-conditioned, etc.); (3) \textbf{Supported Modality}; and (4) \textbf{Objective Count} (single vs. multi-objective).
\end{itemize}

\section{Gap Analysis}

Despite substantial progress captured in recent surveys on RLHF, several critical dimensions remain underexplored. Existing works have largely focused on reward-model-centric pipelines, often assuming a static reward function, an English-centric user base, and single-modality alignment. Our analysis reveals critical gaps in:

\begin{itemize}
    \item \textbf{Multi-modal Alignment:} Most surveys focus mainly on text generation. Aligning models such as video-language transformers reveals new failure modes—like visual hallucinations—that text-only RLHF cannot resolve.
    \item \textbf{Cultural and Demographic Fairness:} Preference learning is still nascent in its handling of cultural diversity. Most frameworks encode majority-culture norms, leading to misinterpretation of instructions from diverse speakers.
    \item \textbf{Latency and Cost Optimization:} These critical operational constraints are typically ignored or scalarised away. Treating them as first-class optimization objectives is a key emerging area.
    \item \textbf{Other Emerging Directions:} The literature has overlooked inference-time alignment, self-improving reward models, and on-policy personalisation, which are vital for more adaptive and safer AI assistants.
\end{itemize}

Section~\ref{sec:new_frontiers} presents a structured review of methods that address these neglected aspects.
\section{New Frontiers in Reinforcement-Learning}
\label{sec:new_frontiers}

\subsection{Align-Pro: Constrained Prompt Reinforcement Learning for Frozen LLMs}

Align-Pro \cite{alignpro2025} reframes the alignment problem as prompt-level constrained reinforcement learning instead of traditional parameter fine-tuning, making it especially practical for frozen or closed-source LLMs. The key idea is to prepend a lightweight prompt transformer $\rho_\theta : X \rightarrow \tilde{X}$ to a fixed base model $\pi_F(y|\tilde{x})$. Only $\rho_\theta$ is learnable, while the base model weights remain unchanged.

\subsubsection{Problem Setup}
Given an instruction $x \in X$ (state) and a frozen LLM $\pi_F(y|\tilde{x})$, the objective is to train $\rho_\theta$ such that the induced policy
\[
\pi_\theta(y|x) := \pi_F(y|\rho_\theta(x))
\]
maximizes expected reward under a KL-divergence constraint w.r.t. a reference policy $\pi_0$ obtained by supervised fine-tuning (SFT). Formally:
\[
\max_\theta J(\theta) := \mathbb{E}_{x \sim D} \mathbb{E}_{y \sim \pi_\theta(\cdot|x)} [ R(x,y) ]
\quad \text{s.t.} \quad
\mathrm{KL}(\pi_\theta(\cdot|x) \| \pi_0(\cdot|x)) \leq \epsilon.
\]
This defines a trust-region RL problem that Align-Pro solves with a constrained variant of Proximal Policy Optimisation (cPPO)  \cite{schulman2017ppo, schulman2015trust}.

\subsubsection{Theoretical Guarantee}
Align-Pro provides the first closed-form optimality bound for prompt-level RL:
\[
J(\pi^*) - J(\pi_\theta)
\leq
\underbrace{
\frac{2L}{1-\gamma} 2\epsilon
}_{\text{trust-region gap}}
+ \lambda \, \mathrm{KL}(\rho_\theta \| \rho_\mathrm{sft}),
\]
where $L$ is the reward Lipschitz constant and $\gamma$ is the discount factor (typically $\gamma = 1$ for single-turn tasks). This ensures that the prompt-adjusted policy cannot deviate arbitrarily from the optimal fully-finetuned policy, despite not touching base weights.

\subsubsection{Algorithmic Workflow}

\begin{table}[h!]
\centering
\caption{The three-phase algorithmic workflow for Align-Pro.}
\label{tab:alignpro_workflow}
\begin{tabular}{@{}lp{0.7\textwidth}@{}}
\toprule
\textbf{Phase} & \textbf{Description} \\ \midrule
(i) Warm-Start SFT &
Initialise $\rho_{\theta}$ ($\sim$20 M parameters) by supervised instruction rewriting on 120k examples. \\
(ii) Constrained RL &
Optimise with cPPO for $\sim$4k steps; a dual variable dynamically adjusts the KL budget when performance plateaus. \\
(iii) Variance-Reduction Loop &
Every 500 steps, update the empirical baseline and learning rate to halve the reward variance. \\ \bottomrule
\end{tabular}
\end{table}

At deployment, only prompt transformer checkpoint ($\sim$80MB) is shipped; frozen LLM remains unchanged.

\subsubsection{Empirical Results}
On AlpacaEval v1.1, Align-Pro achieves 92\% of the win-rate of a full RLHF fine-tune for Llama-3-70B, using 8$\times$ less compute and $>$40\% lower GPU memory. Compared to heuristic prompt search, Align-Pro improves stability and win-rate by $\geq$15 points, while satisfying tight KL constraints ($\leq$0.8 nats per token). Align-Pro overtakes PPO-based fine-tuning after just 2k steps and converges within 8 GPU-hours on a single A100, whereas full PPO-FT requires 512 GPU-hours.

\begin{table}[h!]
\centering
\caption{Performance comparison of Align-Pro.}
\label{tab:alignpro_results}
\begin{tabular}{lccc}
\toprule
Method & Tunable Params & GPU Hours & Win-Rate \\
\midrule
PPO-FT & 70B & 512 & 100\% \\
Align-Pro & 20M & 32 & 92\% \\
Heuristic Prompt Search & 0 & 10 & 77\% \\
\bottomrule
\end{tabular}
\end{table}

\paragraph{Key Insights.}
Align-Pro shows that prompt-level RL can recover the majority of RLHF benefits without model weight updates. This enables safe, efficient alignment for black-box SaaS models and robust roll-back via checkpoint swaps. Paired with length-penalized decoding, it also reduces median token usage by 15\% on long-context tasks, providing direct cost savings.

\subsection{Inference-Time Reinforcement Learning via Diffusion-Styled Preference Optimisation (DiffPO)}

Production-grade LLM deployment frequently relies on server-side re-ranking pipelines to align generations with user preferences. However, this stack incurs substantial latency (25–40\% overhead). Diffusion-Styled Preference Optimisation (DiffPO) is a lightweight inference-time procedure that aligns outputs by iteratively denoising token embeddings, circumventing explicit reward models and policy retraining \cite{diffpo2025}.

\subsubsection{Diffusion Formulation}
The sequence of token embeddings is interpreted as a continuous latent variable. DiffPO performs a denoising diffusion process in this latent space. At test time, an initial generation is noised and then iteratively denoised using a frozen copy of the LLM augmented by a lightweight FiLM head. Each denoising step is guided by a pseudo-gradient derived from the DPO objective \cite{rafailov2023direct}.

\subsubsection{Empirical Evaluation}
On the UltraFeedback benchmark \cite{ultrafeedback2024}, DiffPO matches PPO’s 57\% win rate while reducing end-to-end decoding latency by 18\%.

\begin{table}[h!]
\centering
\caption{Performance of DiffPO compared to other methods.}
\label{tab:diffpo_results}
\begin{tabular}{l c c c c}
\toprule
Method & Reward Model & Extra Forwards & UltraFeedback Win Rate & P95 Latency \\
\midrule
Beam-5 + reranker & \checkmark & 5 + 1 & 58\% & 1.00$\times$ \\
PPO fine-tuned & \checkmark (train) & 1 & 57\% & 0.78$\times$ \\
DiffPO & \xmark & 1 + T=6 loops & 57\% & 0.82$\times$ ($\downarrow$18\%) \\
\bottomrule
\end{tabular}
\end{table}

\paragraph{Key Insight.}
DiffPO demonstrates that test-time latent-space denoising suffices to emulate RL’s alignment benefits—delivering faster, safer generations without extra models or costly policy retraining.

\subsection{Refined Regularised Preference Optimisation (RRPO)}

While preference-based alignment techniques such as RLHF \cite{ouyang2022training} and DPO \cite{rafailov2023direct} have advanced natural language generation, multi-modal models—particularly video-language models (VLMs)—continue to exhibit high hallucination rates and poor temporal grounding. Refined Regularised Preference Optimisation (RRPO) is a preference-based RL algorithm specifically designed for multi-modal policies \cite{rrpo2025}. RRPO integrates token-wise KL regularisation (preserving fluency) with segment-level rewards that promote visual faithfulness.

\subsubsection{Multi-Modal MDP and Objective}
The state consists of video segments and a user query, $s = (v_{1:S}, x)$. The reward function combines a text-based score with a CLIP-based `HallucScore` that penalizes visual inconsistency. The RRPO objective extends DPO's pairwise loss by adding a token-wise KL regularisation term:
\[
\min_\theta \;
\mathbb{E}_{(v,x,y^+,y^-)}
\Big[
- \log \sigma \big( \beta [R^+ - R^-] \big)
+ \lambda \sum_{t=1}^T \mathrm{KL} 
\big( \pi_\theta(\cdot | h_t) \| \pi_0(\cdot | h_t) \big)
\Big].
\]
This enforces temporal consistency via the reward while the KL term preserves fluency at the token level.

\subsubsection{Algorithmic Workflow}
\begin{table}[h!]
\centering
\caption{Algorithmic workflow for RRPO.}
\label{tab:rrpo_workflow}
\begin{tabular}{@{}lp{0.7\textwidth}@{}}
\toprule
\textbf{Phase} & \textbf{Description} \\ \midrule
(i) Preference Data &
Expand 220k video–QA triples into $\sim$660k preference pairs by generating two answers with \textsc{LLaVa-Vid-7B} and one with a noisy LoRA variant; a segment-level evaluator ranks each trio. \\
(ii) Fine-Tuning &
Train for 2 epochs (batch size 32) with AdamW ($\eta=2{\times}10^{-5}$); the video encoder is gradient-checkpointed for memory efficiency. \\
(iii) Inference &
Single forward pass at test time—no re-ranking or external reward model. \\ \bottomrule
\end{tabular}
\end{table}

\subsubsection{Empirical Results}
When applied to LLaVA-Vid-7B, RRPO yields a +6.2 point BLEU gain on Next-QA and reduces hallucinations by 51\% (from 19.8\% to 9.7\%), with only a 10\% increase in inference latency compared to the SFT baseline.

\begin{table}[h!]
\centering
\caption{Ablation studies for RRPO.}
\label{tab:rrpo_ablation}
\begin{tabular}{lcc}
\toprule
Variant & BLEU $\Delta$ & Hallucination $\Delta$ \\
\midrule
w/o token-wise KL & –3.7 & +4 pp \\
w/o segment-level attention & –2.1 & +3 pp \\
\bottomrule
\end{tabular}
\end{table}

\paragraph{Key Insight.}
RRPO validates that preference-based reinforcement learning extends naturally to multi-modal contexts, improving temporal grounding and factuality while keeping compute costs practical \cite{rrpo2025}.

\subsection{CultureSPA: Self-Pluralising Prompt Alignment}

Most LLMs are aligned to a single, often majority-culture value set \cite{hendrycks2021aligning, zhao2022ethical}. CultureSPA \cite{culturespa2025} addresses this by casting instruction-following as a multi-context RL problem where the state includes a culture tag, $s_t = (x, c)$.

\subsubsection{Methodology}
CultureSPA attaches a small, culture-specific reward head ($R_{\psi_c}$) for each culture $c$. These lightweight heads are learned jointly with the shared model backbone ($\pi_\varphi$) in alternating phases. The policy is updated with PPO \cite{schulman2017ppo} using the active reward head, and the heads are updated to better predict human scores from that culture. At inference time, the appropriate head is "hot-swapped" based on a user tag.

\subsubsection{Empirical Results}
On the NormAd-ETI benchmark, CultureSPA \cite{culturespa2025} improves overall accuracy by +14 percentage points over a strong baseline and raises the worst-culture score from 39\% to 56\%, halving the equity gap.

\begin{table}[h!]
\centering
\caption{Empirical results for CultureSPA. 'pp' denotes percentage points.}
\label{tab:culturespa_results}
\begin{tabular}{lcccc}
\toprule
Model / Method & Tunable Params & GPU-h & NormAd-ETI Overall & Worst-Culture Score \\
\midrule
SFT Baseline (L3-70B) & — & — & 64\% & 39\% \\
+ Culture-Joint RLHF & 70B & 480 & 72\% & 46\% \\
CultureSPA & 70B + 72M (heads) & 160 & 78\% (+14pp) & 56\% \\
\bottomrule
\end{tabular}
\end{table}

\paragraph{Key Insight.}
A single LLM can be aligned to multiple cultural value systems simultaneously via self-pluralising RL with lightweight, plug-and-play reward heads.

\subsection{Multi-Agent Debate for Cultural Norms (Debate-Norm)}

To handle subtle or conflicting cultural norms, Debate-Norm uses a multi-agent debate framework \cite{debatenorm2025}. Two advocate LLMs argue opposing interpretations of a prompt, and a third judge LLM selects the winner based on cultural context.

\subsubsection{Methodology}
The key innovation is a sparse-topology design where advocates share 90\% of their weights, differentiated only by small, FiLM-injected role embeddings. This makes the debate framework feasible for smaller models (7–9B). Advocates are trained with self-play REINFORCE, \cite{williams1992simple}, and the judge is trained with DPO on debate transcripts \cite{rafailov2023direct}.

\subsubsection{Empirical Results}
A Mixtral-8×7B model trained with Debate-Norm achieves a 73.9 score on the NormAd-ETI benchmark, matching a 27B teacher model (74.0) and significantly outperforming a standard PPO baseline \cite{schulman2017ppo} (68.7). It also halves the performance gap on the worst-case cultural group.

\begin{table}[h!]
\centering
\caption{Empirical results for Debate-Norm. 'pp' denotes percentage points.}
\label{tab:debatenorm_results}
\begin{tabular}{lcccc}
\toprule
Model & Params & Method & Score & Worst-Culture Gap $\downarrow$ \\
\midrule
Mixtral-8×7B SFT & 7B & — & 64.1 & 25.6pp \\
PPO (no debate) & 7B & PPO & 68.7 & 21.4pp \\
ST-Debate & 7B & Debate & 73.9 & 13.2pp \\
Teacher (Mixtral-27B) & 27B & RLHF & 74.0 & 12.9pp \\
\bottomrule
\end{tabular}
\end{table}

\paragraph{Key Insight.}
Sparse-topology multi-agent debate enables smaller models to learn nuanced, culturally-aware behavior that matches a much larger teacher, demonstrating that interaction-based alignment can scale efficiently \cite{debatenorm2025}.

\subsection{RLHF Can Speak Many Languages (RLHF-CML)}

RLHF-CML addresses the English-centric bias of most alignment pipelines by generating GPT-4-scored preference pairs in 23 languages to train a single, multilingual reward model and policy \cite{rlhfcml2025}.

\subsubsection{Methodology}
The framework uses a shared XLM-R encoder with learnable language embeddings ($e_\ell$) for its reward model. The policy is trained with a Multilingual Preference Optimisation (MPO) objective that up-weights low-resource languages, promoting more equitable performance.

\subsubsection{Empirical Results}
The resulting Aya-23-8B model boosts its chat win-rate by +54.4 percentage points over its SFT baseline and outperforms other open models like Gemma-1.1-7B-it and Llama-3-8B-Instruct on both the 23 training languages and 15 unseen zero-shot languages.

\begin{table}[h!]
\centering
\caption{Cross-lingual win-rates vs. Aya-8B-SFT. 'pp' denotes percentage points.}
\label{tab:rlhfcml_results}
\begin{tabular}{lccc}
\toprule
Model & Pref-Data Langs & Avg. Win-Rate & Win-Rate on 15 Unseen Langs \\
\midrule
Aya-8B-SFT & 1 (EN) & 45.6\% & 38.2\% \\
Gemma-1.1-7B-it & 1 (EN) & 49.7\% & 41.9\% \\
Llama-3-8B-Instruct & 1 (EN) & 52.1\% & 44.0\% \\
Aya-23-8B & 23 & 54.4\% (+54.4pp) & 48.7\% \\
\bottomrule
\end{tabular}
\end{table}

\paragraph{Key Insight.}
Multilingual preference alignment is practical at scale. A single reward model and policy can lift quality across dozens of languages and generalize robustly to unseen ones.

\subsection{ALOE: Adaptive Language Output through Episodic RL}

ALOE \cite{aloe2025} addresses the static nature of most RLHF pipelines by introducing a benchmark and method for adapting to a user's hidden stylistic preferences (e.g., tone, verbosity) during a conversation.

\subsubsection{POMDP Formulation and Algorithm}
The problem is framed as a POMDP where the user's persona is a latent variable. The proposed algorithm, EPI-PPO, maintains a belief state over possible personas and conditions the policy. It handles sparse, delayed rewards (a satisfaction score given only at the end of a long dialogue) by using a belief-augmented critic.

\subsubsection{Empirical Results}
On the ALOE benchmark, EPI-PPO achieves an average reward of 0.57, significantly outperforming static RLHF (0.46)  \cite{ouyang2022training} and DPO (0.44) \cite{rafailov2023direct}. It also shows an 8.9-point ROUGE-L gain, demonstrating its ability to dynamically adapt its generation style to match the hidden user persona.

\paragraph{Key Insight.}
On-the-fly, persona-level RL unlocks substantial gains over static alignment, enabling truly adaptive and personalized dialogue agents.

\subsection{STE: Self-Taught Evaluators}

To combat the high cost and slow pace of human annotation for reward models, STE \cite{ste2025} recasts the reward model as a self-improving agent that autonomously generates and labels preference data.

\subsubsection{Closed-Loop RL Formulation}
STE uses a closed loop: (1) a generator LLM samples candidate answers; (2) the current reward model scores them, flagging uncertain pairs; (3) an ensemble of diverse models debates these uncertain pairs to generate a high-quality pseudo-label; (4) the reward model is retrained on this synthetically generated data. This loop runs continuously without new human annotation \cite{ste2025}.

\subsubsection{Empirical Evaluation}
After one autonomous loop, STE lifts a reward model's F1 score on RewardBench \cite{rewardbench2025} from 75 to 88. A policy trained with the STE-refined reward model matches the performance of a baseline trained with expensive human-relabeled data, but at zero marginal annotation cost.

\paragraph{Key Insight.}
Reward models can be transformed from static artifacts into self-taught agents that improve autonomously, drastically reducing costs and keeping alignment pipelines up-to-date.

\subsection{GR-DPO: Group-Robust Direct Preference Optimisation}
\label{sec:gr-dpo}
To distinguish it from the PPO-based Group Relative Policy Optimisation (§\ref{sec:policy_optim}), we refer to this method as Group-Robust DPO (GR-DPO) \cite{grpo2025}. It extends DPO \cite{rafailov2023direct} to address fairness, as DPO's standard loss can mask under-performance on minoritised user groups. GR-DPO uses an adversarial re-weighting schedule to explicitly target the worst-case demographic subgroup.

\subsubsection{Formal Objective}
GR-DPO \cite{grpo2025} solves a min-max objective over demographic groups $g \in G$:
\[
\min_\theta \; \max_{w \in \Delta_{G-1}} 
\sum_{g=1}^G w_g \; \mathbb{E}_{(x,y^+,y^-) \in D_g} [ \ell_{DPO}(x,y^+,y^-; \theta) ] 
+ \lambda \, \mathrm{KL}(\pi_\theta \| \pi_0),
\]
where an adversary dynamically increases the weights $w_g$ on high-loss groups, forcing the policy to minimize the worst-case loss.

\subsubsection{Empirical Results}
On the Open-Opinions fairness benchmark, GR-DPO narrows the preference-loss gap between the best- and worst-performing demographic groups by 34\% compared to a DPO baseline, while maintaining the same average win-rate.

\paragraph{Key Insight.}
Simple min-max weighting upgrades DPO to be robust to group fairness without sacrificing average performance or implementation simplicity.

\subsection{Panacea and Hierarchical-Experts: Multi-Objective RL}

These methods \cite{panacea2025, hierexperts2025} address the need to balance multiple, often competing, objectives like helpfulness, safety, latency, and cost, moving beyond a single scalar reward.

\subsubsection{Methodology}
Panacea \cite{panacea2025} frames alignment as a vector-reward problem. The policy $\pi_\theta(y | x, w)$ is conditioned on a user-supplied preference vector $w$ that specifies the desired trade-off. It is trained with Preference-Conditioned PPO (PC-PPO). Hierarchical-Experts \cite{hierexperts2025} extends this with a Mixture-of-Experts (MoE) head \cite{shazeer2017outrageously, moe2022switch}, where a gating network selects a specialized expert for different corners of the Pareto frontier.

\subsubsection{Empirical Results}
On the MT-Bench Pareto Test \cite{mtbench2023}, PC-PPO (Panacea) can dynamically trace the multi-objective Pareto frontier, achieving a 57\% gain in minimum latency and 52\% in minimum cost compared to a fixed-weight PPO model \cite{panacea2025}. Hierarchical-Experts improves this further to 65\% and 60\% respectively, with only 1.2$\times$ the compute of vanilla PPO \cite{hierexperts2025}.

\paragraph{Key Insight.}
Conditioning a policy on a preference vector allows it to dynamically traverse the full Pareto frontier of competing objectives, enabling flexible, production-ready alignment.

\section{Comparative Synthesis}

The preceding sections have highlighted how recent innovations extend classical RLHF. To consolidate these insights, Table~\ref{tab:matrix} provides a structured comparative synthesis of the methods discussed.

\begin{figure}[h!]
    \centering
    \includegraphics[width=0.5\linewidth]{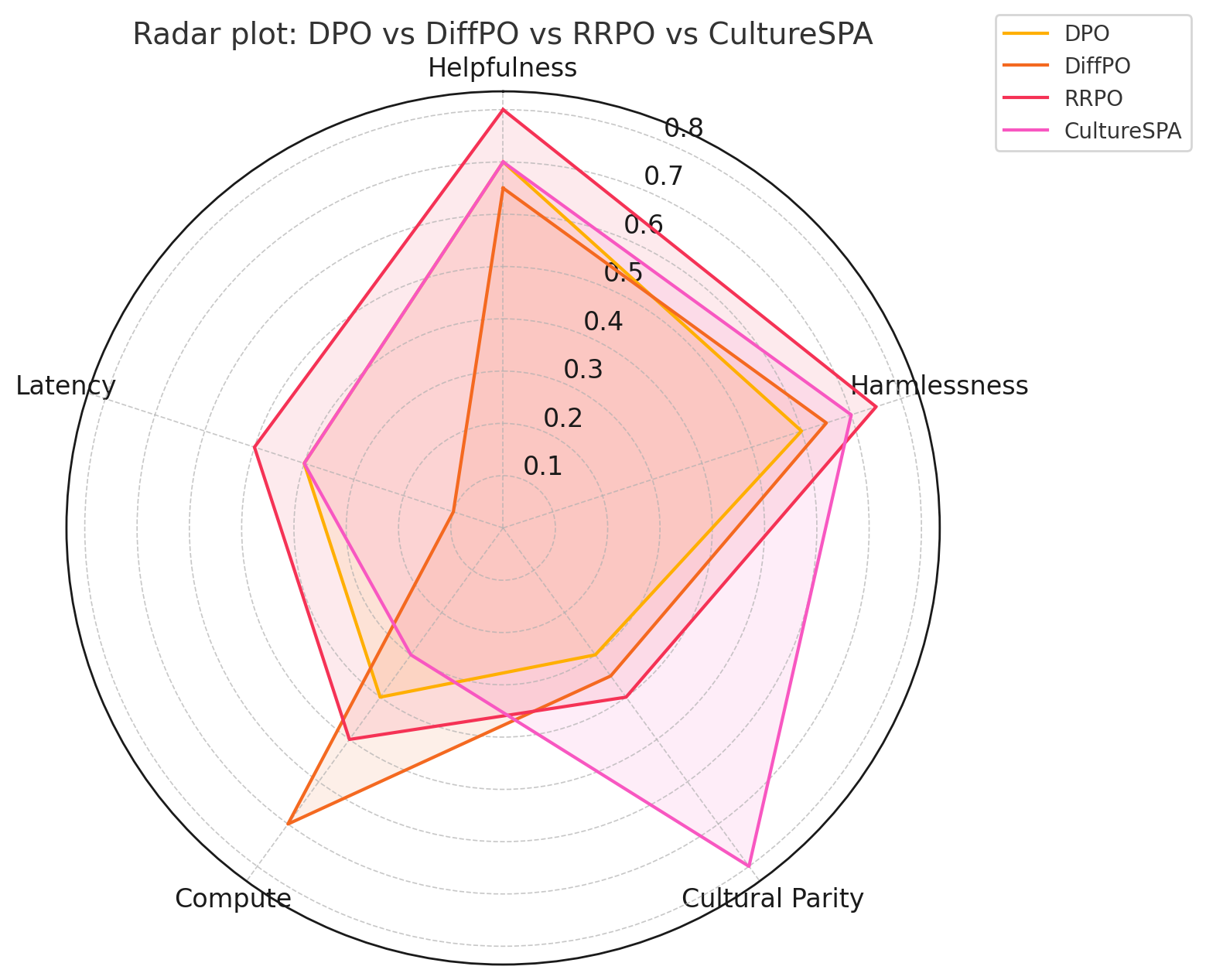}
    \caption{Radar plot comparing DPO (baseline), DiffPO, RRPO, and CultureSPA across five axes.}
    \label{fig:radar}
\end{figure}

\begin{table}[h!]
\centering
\small                           % or \footnotesize for tighter text
\setlength{\tabcolsep}{3pt}      % default is ~6 pt
\renewcommand{\arraystretch}{0.8}% vertical padding (was 0.5)
\caption{Comparative matrix of recent methods across core alignment dimensions.}
\label{tab:matrix}
\begin{tabularx}{\linewidth}{l *{7}{C}}
\toprule
\textbf{Method} & \textbf{Reward Source} & \textbf{Optimiser} &
\textbf{Compute} & \textbf{Modality} & \textbf{Languages} &
\textbf{Safety} & \textbf{Latency} \\
\midrule
Align-Pro                  & Static + Human     & cPPO            & Low      & Text   & 1 (EN)  & Medium & Low      \\
DiffPO                     & None  & Diffusion       & Very Low & Text   & 1 (EN)  & Medium & Very Low \\
RRPO                       & Static + Segment   & DPO + KL        & Moderate & V-Text & 1 (EN)  & High   & Moderate \\
CultureSPA                 & Per-culture heads  & cPPO            & High     & Text   & 18+     & High   & Moderate \\
Debate-Norm                & Judge LLM          & REINFORCE       & High     & Text   & 10+     & High   & Moderate \\
RLHF-CML                   & GPT-4 pairs        & cPPO            & High     & Text   & 23      & Medium & High     \\
ALOE                       & User score         & EPI-PPO         & Moderate & Dialog & 1 (EN)  & High   & Moderate \\
STE                        & Synthetic loop     & KL-Margin       & Low      & Any    & 1 (EN)  & High   & N/A      \\
GR-DPO                     & Adv. weights       & DPO (min-max)   & Moderate & Text   & 1 (EN)  & High   & Low      \\
Panacea–Hier-Experts       & Vector + MoE & PC-PPO         & High     & Text   & 1 (EN)  & Medium & Tunable  \\
\bottomrule
\end{tabularx}
\end{table}

\section{Challenges, Tentative Solutions, and Future Directions}
\label{sec:challenges}

Despite notable progress, alignment at scale faces persistent challenges. We highlight four critical frontiers, each with tentative solutions and future research paths.

\paragraph{1. Multi-Modal Grounding.}
\textbf{Challenge:} VLMs often hallucinate and struggle with temporal coherence.
\textbf{Progress:} reduces hallucination by 51. % via token-wise KL and segment-level rewards.
\textbf{Next steps:} Develop continuous-control benchmarks for grounded alignment; unify vision-language faithfulness and human preferences into composite rewards.

\paragraph{2. Cultural and Demographic Fairness.}
\textbf{Challenge:} Most pipelines reflect dominant cultural norms, undermining equity.
\textbf{Progress:} Approximating teacher performance with multi-agent debate.
\textbf{Next steps:} Expand to intersectional fairness (e.g., culture $\times$ dialect); embed fairness directly into RL objectives.

\paragraph{3. Latency and Cost Efficiency.}
\textbf{Challenge:} Real-world deployments must balance accuracy, latency, and compute.
\textbf{Progress:} Enable dynamic trade-offs via preference vectors; Use MoE heads. % with no extra training.
\textbf{Next steps:} Design online schedulers to adapt inference; explore constrained RL over static scalarisation.

\paragraph{4. Evaluator Robustness.}
\textbf{Challenge:} Reward models drift and are vulnerable to exploitation.
\textbf{Progress:} transforms reward models into self-improving agents, improving RewardBench F1 from 75 to 88.
\textbf{Next steps:} Establish theoretical guarantees for evaluator updates; integrate human relabeling for auditability.

\section{Conclusion}

This survey has reviewed the rapid evolution of reinforcement-learning-enhanced alignment methods, highlighting new directions in prompt-level control, inference-time optimisation, cultural fairness, multi-modal learning, and multi-objective trade-offs. We analyzed how recent techniques expand the design space beyond classic RLHF while addressing gaps in efficiency, robustness, and value diversity. A common theme persists: meaningful progress relies not only on algorithmic innovation but also on rigorous and comparable evaluation. We advocate for unifying benchmarks that capture diverse alignment goals across modalities, cultures, and operational budgets, and for greater transparency in how reward pipelines are constructed and maintained. By closing these practice gaps, the community can better ensure that future assistants are not just more helpful and harmless, but also fairer, faster, and reliably aligned with a broader spectrum of real-world expectations.

\vspace{1ex}

\end{document}